\documentclass[12pt]{article}
\usepackage[utf8]{inputenc}
\usepackage{amssymb, amsmath, amsthm}
\usepackage[pdftex]{graphicx}
\usepackage{hyperref}
\usepackage{color}
\usepackage{float}
\usepackage{lineno}
\usepackage{geometry}
\begin{document}

\begin{titlepage}
\vspace{0.4cm}

\setlength{\baselineskip}{0.6cm}

\begin{center}

{\Large {\bf
Transfer entropy-based feedback improves performance in artificial neural networks}}

\vspace{0.4cm}

{\bf
Sebastian Herzog, Christian Tetzlaff and Florentin W\"org\"otter*}

\vspace{0.3cm}

\normalsize
\noindent
Universit\"at G\"ottingen,\\ Bernstein Center for Computational Neuroscience,\\ III Physikalisches Institut - Biophysik,\\ G{\"o}ttingen, Germany
\end{center}

\vskip 2cm
\begin{abstract}
The structure of the majority of modern deep neural networks is characterized by unidirectional feed-forward connectivity across a very large number of layers. By contrast, the architecture of the cortex of vertebrates contains fewer hierarchical levels but many recurrent and feedback connections. Here we show that a small, few-layer artificial neural network that employs feedback will reach top level performance on a standard benchmark task, otherwise only obtained by large feed-forward structures.
To achieve this we use feed-forward transfer entropy between neurons to structure feedback connectivity. Transfer entropy can here intuitively be understood as a measure for the relevance of certain pathways in the network, which are then amplified by feedback. Feedback may therefore be key for high network performance in small brain-like architectures.
\end{abstract}

\vskip 7cm
\noindent
*Correspondence:\\ Florentin W\"org\"otter \\
Universit\"at G\"ottingen,\\ III Physikalisches Institut - Biophysik, Friedrich-Hund Platz 1,\\ DE-37077 G{\"o}ttingen, Germany\\
Email: worgott@gwdg.de

\end{titlepage}

\newpage

Modern deep neural networks employ as many as 152 hierarchical layers \cite{inceptionv4,resnet} between input and output, whereas vertebrate brains achieve high levels of performance using a much shallower hierarchy. This may well be largely due to massive  recurrent and feedback connections, which are dominant constituents of cortical connectivity \cite{Markov2014}. Their role remains puzzling in artificial neural networks.

Intra-layer recurrent connections have indeed become an important aspect in several deep learning architectures, notably deep recurrent neural nets (DRNNs) \cite{Hermans2013,Liao2016} and especially also in  LSTM-networks \cite{hochreiter1997long}, which show superior performance as compared to conventional DRNNs. Different from this, inter-layer feedback is less common and mostly employed in rather specific ways. For example, feedback connections have been introduced into deep Boltzmann Machines, which are an unsupervised method \cite{Salakhutdinov2009}. Alternatively, feedback has been used in deep architectures to create the equivalent of selective attention \cite{Stollenga2014,Cao2015} or to implement some aspect of long-term memory for object recognition \cite{Varadarajan2013}. Several other approaches exist that employ feedback in deep learning architectures in different ways and for different applications \cite{Oberweger2015,Veeriah2015,Shi2016,Testolin2016}, including connections that only influence learning \cite{Lillicrap2016}. Thus, the literature on the use of feedback in artificial neural networks appears diverse (for general reviews see also \cite{Lecun2015,Roudi2015}).

All in all, the architectural differences between real and artificial neural networks (with or without feedback) make it difficult to compare deep neural networks and their performance to brain structure and function. Several relations between both substrates have been reported \cite{Khaligh2014,Guclu2015,Kriegeskorte2015,Cichy2016,Yamins2016}
but processing architectures differ too much to allow for direct comparison. Differences and potential relations, however, are currently vividly discussed \cite{Bengio2015,Liao2016,Marblestone2016}.

One central problem, which has so far prevented widespread use of feedback in artificial nets, is how to actually structure feedback connectivity. Feedback had been investigated in great detail for the visual system where it can influence spatial- as well as object- and feature-oriented attention. It can affect perceptual tasks and object expectation, and it can also help to create efference copies, influence perceptual learning, and guide different aspects of memory function (reviewed in \cite{Gilbert2013,Spillmann2015}).

One dominant aspect of feedback, potentially common to all these influences, is to amplify a feed-forward processing pathway and increase its signal throughput. Thus, we asked whether we could positively influence the performance of an artificial neural network using a similar mechanism?

We were especially interested in using here a rather small deep learning net (AlexNet, \cite{alexnet}) with few layers, which --- as compared to the state-of-the-art --- performs sub-standardly, asking how far image classification performance can be improved by feedback. To define feedback and to enhance signal throughput, we use the transfer entropy between network nodes. Previously this measure has been used for estimation of functional connectivity of neurons \cite{Lizier2011,Vincente2011,Shimono2015}. Thus, it offers a natural gauge for adjusting feedback connectivity. This way we add about 75\% more connections without additional training to the network, all of which, however, are very small. As a result this only mildly amplifies already connected paths. Nonetheless, we now gain --- robustly against several controls --- 10\% improvement on a common benchmark problem, thereby reaching the top-performing group of state-of-the-art networks.

\begin{figure}[H]
 \centering
 \includegraphics[width=1.0\linewidth]{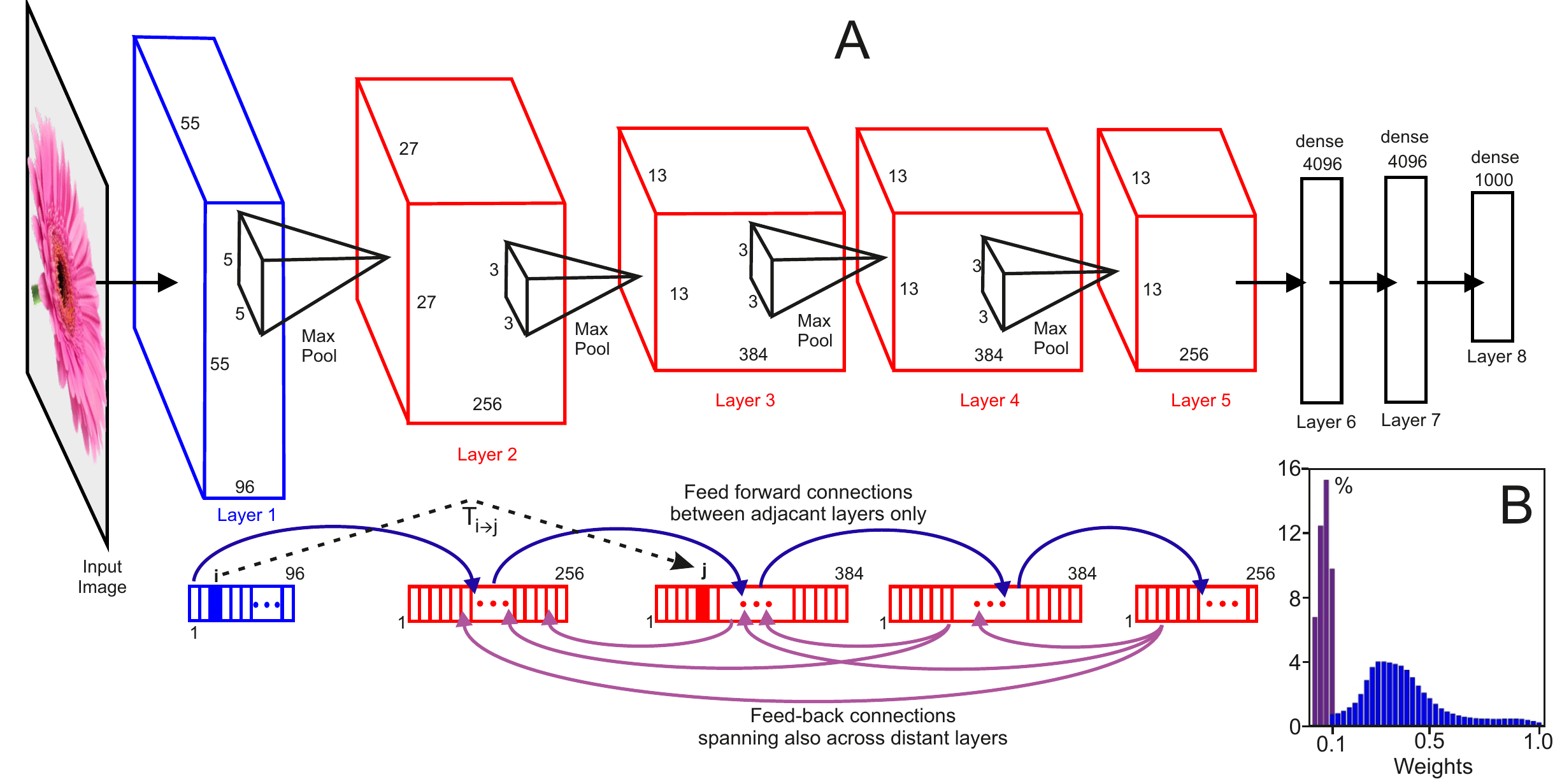}
 \caption{\emph{\textbf{(A)} Architecture of the AlexNet \cite{alexnet} and its feed forward and feedback structure. Layers 1 to 5 perform convolutions, layers 6 to 8 perform the final scoring for classification. Numbers 96, 256, 384, 1000, and 4096 refer to numbers of nodes in the different layers. Numbers near the pyramid-icons refer to the max-pooling kernel sizes. The other numbers represent the sizes of the convolution kernels. \textbf{(B)} Weight distribution (in percent) of the FF- and FB-AlexNet. Feedback weights in purple.}}
 \label{fig:Alex}
\end{figure}

\section*{Results}
\noindent
\underline{\textbf{Data and Network:}} We focus on an image classification task for which we use the labeled CIFAR-10 data set $\mathcal{X}= \{\mathcal{X}_1, \mathcal{X}_1,\cdots ,\mathcal{X}_{10}\}$, with 10 image classes $\mathcal{X}_q$. In total, CIFAR-10 contains 60,000 color images and it is a standard benchmark \cite{Krizhevsky09learningmultiple,Krizhevsky2009}.

The AlexNet \cite{alexnet} has eight layers. Its core consists of five layers, which perform convolutions. Only layers 2-5 will form feedback connections; the input layer (Layer 1) and the three dense layers (Layers 6-8) do not. Layers 2-5 consist of 1280 neurons (nodes) distributed as shown in figure~\ref{fig:Alex}. Initially, all feed-forward network weights $w$  are determined by the network training process.

Activation of network nodes $i$ is defined as $g(x_i)$, where $x_i$ is the output of the belonging convolution kernel and $g$ is a sigmoidal activation function $g(x)=ln(1+exp(x))$, which keeps network activity bounded.\\

\noindent
\underline{\textbf{Process:}} The goal is to show how certain feedback connections will improve image classification performance. The whole process takes five steps.
\begin{enumerate}
\item
  Train the weights for the feed-forward AlexNet (\emph{FF-AlexNet}) using a subset of the CIFAR data, employing standard error back-propagation training \cite{Rumelhart1986,Schmidhuber14}. This generates all feed-forward network weights $w$. Then, determine the smallest weight $w_{min}$ in the whole FF-AlexNet. This is later the maximum weight for any feedback connection.
\item
  Run the FF-AlexNet on the complete CIFAR data. Store all activations $g_i$ for all nodes $i$. In addition, for the calculation of the transfer entropies, determine and store \emph{activation events} $y_i$. Given that networks of this size most often show weak activations, we want to register as an event \emph{any} activation above a small numerical inaccuracy limit. Thus, we set $y_i=1$ if $g_i > 0.001$ and zero otherwise. This way the transfer entropy can be calculated considering all (even weakly active) nodes.
\item
  Take all images from an image class $\mathcal{X}_q$ that have the same label and determine the transfer entropies $T$ for every node $i$ to all its direct \emph{and} indirect target nodes $j$:  $T^{\mathcal{X}_q}_{i\rightarrow j}$. ''Indirect'' here means nodes that are not connected to the source node, but that receive inputs from it upstream through a connected pathway. Naturally, nodes $i$ and $j$ will always belong to different layers as there are no lateral connections. Note that transfer entropies consider also the input layer (see example in Fig.\ref{fig:Alex}), which, however, will not receive feedback connections. Calculation of the transfer entropies (Eqs.~\ref{eqn:te1}-\ref{eqn:te3}) requires using activation events $y_i, y_j$ as defined in the previous step. Repeat this for all image classes and average, getting $\tilde {T}_{i\rightarrow j}={{1}\over{10}}\sum_{q=1}^{10}  T^{\mathcal{X}_q}_{i\rightarrow j}$.
\item
  Then define feedback $f$ from node $j$ to $i$ by:
  \begin{equation}
  f_{{j_\beta} \rightarrow {i_\alpha}}=
  \begin{cases}
  \frac{w_{min} |\beta -\alpha|}{L} & \text{for } \tilde {T}_{i_\alpha\rightarrow j_\beta} < \Phi\\
  0, & \text{else}
  \end{cases}
  \label{eqn:fb}
  \end{equation}
  Here $\beta > \alpha$ are layer numbers, and $L=4$, the number of maximally considered layers. Thus, feedback is larger than zero only if the mean transfer entropy between the two nodes $\tilde {T}_{i_\alpha\rightarrow j_\beta}$ remains below threshold $\Phi$. Note Eq.~\ref{eqn:fb} amplifies far-reaching feedback paths on average more than paths between neighboring layers, which is an essential feature of this system (see "Discussion").
\item
  Run the feedback AlexNet (\emph{FB-AlexNet}) on the whole data set, where for the feed-forward-feedback activation $\tilde{g}(x_i)$ we define: $\tilde{g}(x_i) = g\left(x_i + \sum_j {f_{{j} \rightarrow i}}\right)$ summing over all nodes $j$ that provide feedback to $i$.
\end{enumerate}
After the last step we have obtained the final classification results.\\

\begin{table}[h!]
\centering
\caption{Performance compared to other methods.}
{%
\newcommand{\mc}[5]{\multicolumn{#1}{#2}{#3}{#4}{#5}}
\begin{center}
\begin{tabular}{|c|c|c|c|c|c|c|}\hline
{``Fractional} & {``Striving} & {FB-AlexNet} & {``All you need} & {FF-AlexNet}\\
{Max-Pooling''} \cite{Graham14a} & { for simplicity''} \cite{springenberg2014striving} & & {is a good init''} \cite{MishkinM15} & \\\hline\hline
96.53\% & 95.59\% & 94.62\% & 94.16 \% & 85.50 \%\\\hline
\end{tabular}
\end{center}
\label{tab:compare}
}%
\end{table}

\begin{figure}[h!]
 \centering
 \includegraphics[width=0.6\linewidth]{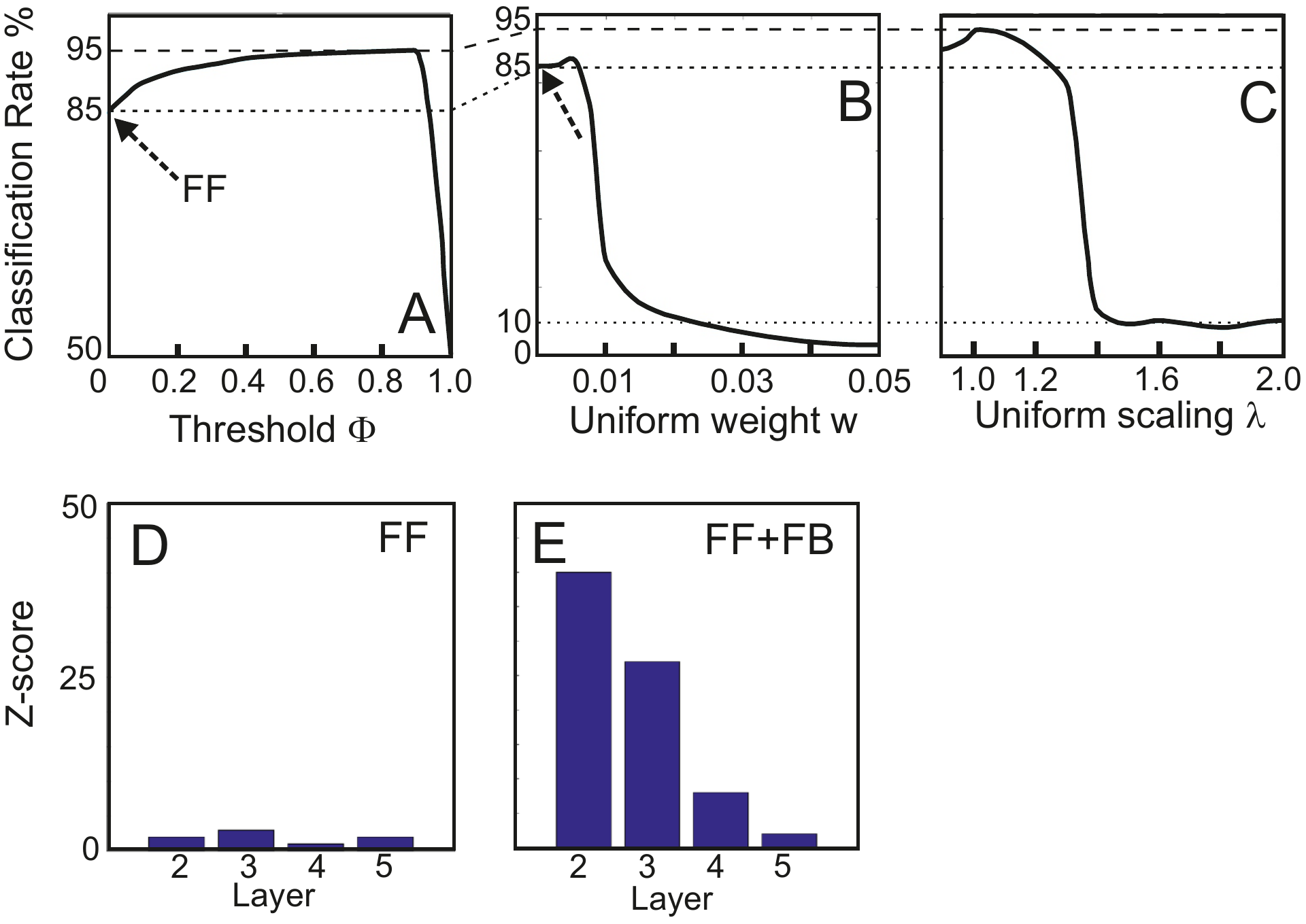}
 \caption{\emph{Performance quantification of the FB-AlexNet and controls. Arrows mark the performance of the FF-AlexNet. \textbf{(A)} Robustness of the feedback against variations of the only control parameter $\Phi$. \textbf{(B)} Control: Network with uniform FB weights. \textbf{(C)} Control: Network with scaled FB weights. \textbf{(D,E)} Local active information storage for the FF- and the FB-AlexNet.}}
 \label{fig:phi}
\end{figure}

\noindent
\underline{\textbf{Quantification:}} The central finding in this study is that the FB-AlexNet improves classification performance by about 10\% against the feed forward network. Without feedback ($\Phi=0$ in Fig.~\ref{fig:phi}~A) the network had a classification rate of 85\% (classification error 14.5\%) and with feedback ($\Phi=0.9$) the network achieves 95\% (classification error 5.38\%). For this we used 50,000 training images and 10,000 test images and the FB-AlexNet reaches the top-performing group \cite{Graham14a,springenberg2014striving,MishkinM15} on the CIFAR data (see Tab.~\ref{tab:compare}). This is to some degree remarkable, because the existing top-performing networks are far more complex (especially consisting of far more layers) than the AlexNet \cite{inceptionv4, resnet} and currently performance improvements are often in the range of less than 1\% when a new architecture is proposed (compare the left two columns in Tab.~\ref{tab:compare}). In addition, the FB-AlexNet uses only one control parameter, $\Phi$, and it is robust against variations (Fig.~\ref{fig:phi}~A). Performance improvement remains near optimal across 40\% of the possible range of $\Phi$ (approx. $0.5<\Phi<0.9$ in Fig.~\ref{fig:phi}~A). The performance then rapidly drops only when too much feedback is added, which is due to massive self-excitation.

\begin{figure}[!h]
 \centering
 \includegraphics[width=0.9\linewidth]{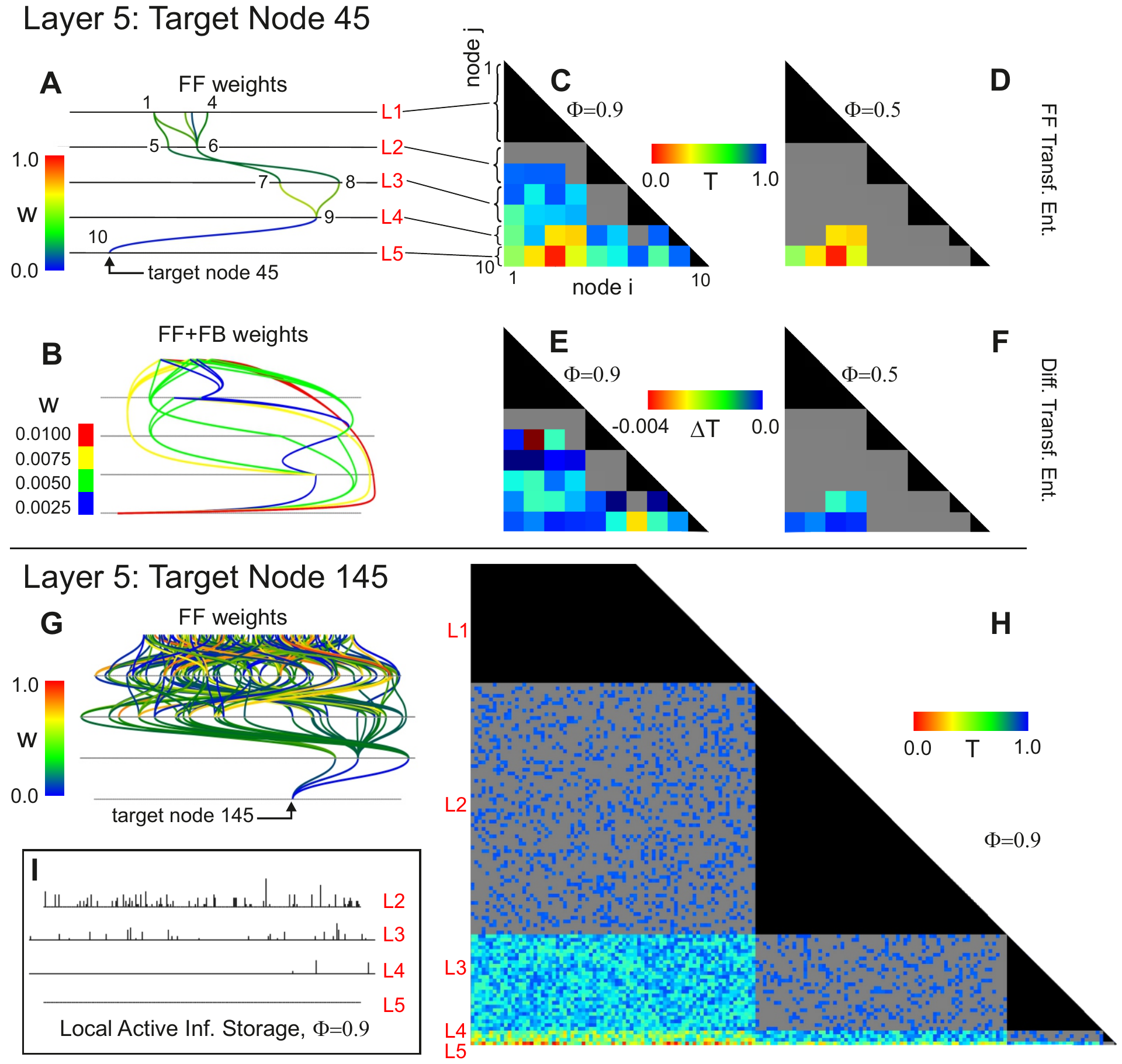}
 \caption{\emph{Detailed analysis of two example convergence trees with layer 5 target nodes as indicated. \textbf{(A)}~Feed-forward weights. Small numbers refer to the consecutive numbering of nodes in diagrams C-F. \textbf{(B)}~Feedback weights (for graphical reasons feed-forward paths are not shown). \textbf{(C,D)}~Transfer entropies $T_{i\rightarrow j}$ for $\Phi=0.9$ (C) and $0.5$ (D) in FF-AlexNet. \textbf{(E,F)}~Difference of transfer entropies without and with feedback, $T^{FF}_{i\rightarrow j}-T^{FF+FB}_{i\rightarrow j}$ for the cases above. \textbf{(G)}~Feed-forward weights for another example path. \textbf{(H)}~Transfer entropies $T_{i\rightarrow j}$ for $\Phi=0.9$. \textbf{(I)} Difference in local active information storage with and without feedback for each node.}}
 \label{fig:TE}
\end{figure}

For $\Phi=0.9$, about 75\% more connections exist in the FB-AlexNet than in the feed-forward network (Table~\ref{tab:eval}) but all have rather small weights (Fig.~\ref{fig:Alex}~B), because feedback weights cannot exceed the smallest existing feed-forward weight $w_{min}$ (Eq.~\ref{eqn:fb}). Interestingly, the introduction of feedback implies a strong, layer-specific increase in the local storage of information as indicated by the local active information storage (Fig.~\ref{fig:phi}~D,E). In general, local active information storage quantifies the amount of information of a unit, which is predictable from the unit's past \cite{wibral2014local}. In feed-forward networks this information is by definition quite low (Fig.~\ref{fig:phi}~D), while feedback connections significantly increase the influence of each unit on its own dynamics (Fig.~\ref{fig:phi}~E and Fig.~\ref{fig:TE}~I). In addition to the increase in local active information storage  feedback connections also adapt other properties of the network. The mean transfer entropy and the characteristic path length \cite{watts1998collective, rubinov2010complex} of the network both drop and, in parallel, the global efficiency \cite{latora2001efficient, rubinov2010complex} increases (Table~\ref{tab:eval}). All these measures taken together indicate an improved processing in the FB-AlexNet.

\begin{table}[b]
\centering
\caption{Different quantification measures for the FF- and FB-AlexNet ($\Phi=0.9$)}.
{%
\newcommand{\mc}[3]{\multicolumn{#1}{#2}{#3}}
\begin{center}
\begin{tabular}{|l||c|c|}\hline
\mc{1}{|c||}{Metric} & \mc{1}{c|}{FF-AlexNet} & \mc{1}{c|}{FB-AlexNet}\\\hline\hline
Total number of nodes & 1280 & 1280\\\hline
Total number of connections & 3877 & 6878\\\hline
Mean transfer entropy & 0.8 & 0.6 \\\hline
Characteristic path length \cite{watts1998collective} & 4.2 & 2.1 \\\hline
Global efficiency \cite{latora2001efficient} & 0.4 & 0.67\\\hline
\end{tabular}
\end{center}
\label{tab:eval}
}%
\end{table}

The widely distributed representation of information in artificial neural networks of this complexity makes it very difficult to find explanations for the here observed performance increase. Still, zooming in on some network nodes provides indications why performance has so strongly improved. Feed-forward connectivity of individual convergence trees (incorporating all paths from the input layer to a specific target unit in the output layer) can vary widely, and two characteristic examples are shown in Figure~\ref{fig:TE}~A,G for layer 5 target neurons 45 and 145. The matrix diagrams in panels C,D, and H display the transfer entropies between nodes $i$ and $j$ along all connected pathways for two different thresholds $\Phi$. Grey regions are those where transfer entropy exceeds $\Phi$ and feedback connections are formed between nodes that have a transfer entropy less than this threshold. Black regions represent intra-layer relations, which are not considered because there are no lateral connections and nodes $i$ and $j$ therefore must belong to different layers.

Figure~\ref{fig:TE}~B shows the feedback for neuron 45 for $\Phi=0.9$ (feed-forward connections not shown). Note that the maximum feedback weight is about a factor of $100$ smaller than the maximum feed-forward weight. As defined by equation~\ref{eqn:fb}, only 4 values for the feedback weights are possible and the largest values are assigned for nodes with biggest layer distances.

Of specific interest is the development of the transfer entropy comparing the pure feed-forward situation with the one that also contains feedback. This is shown in Figure~\ref{fig:TE}~E and F, plotting the difference in transfer entropy: $T^{FF}_{i\rightarrow j}-T^{FF+FB}_{i\rightarrow j}$. In general, feedback connections in the network yield a drop in the transfer entropy by a quite small amount and in a rather dispersed way. This demonstrates that transfer entropy does not just show some overall decrease (Table~\ref{tab:eval}), but, instead some paths are specifically improved by feedback, which are --- presumably --- those that carry most information for classification. A similar picture emerges, when considering local active information storage, shown for target neuron 145 with threshold $\Phi=0.9$ (panel I). Local active information storage is in general increased, but quite differently for different paths.

Several control experiments have been performed to assess the significance of these results. Shuffling of \emph{all} feedback weights or alternatively shuffling of the weights within each pathway renders on average classification rates of only $18.3\pm 7.3$\% or $30.2\pm 10.0$\%, respectively (Table~\ref{tab:control}). Setting all feedback weights $w$ to the same small value produces indeed an initially small improvement to about 87\% (Fig.~\ref{fig:phi}~B), but as soon as weights grow a bit performance drops and, finally, the network tends to massively produce false classifications, leading to less-than-random performance ($<10\%$). Scaling all existing feedback weights (Fig.~\ref{fig:phi}~C) has its maximum at $\lambda = 1.0$, which is our feedback case. For smaller factors the curve approaches 85\%, which is the case without feedback ($\lambda = 0$), for larger factors we reach random performance. Of specific interest is the question of whether an amplification of the feed-forward weights with the weights from the feedback (and then deletion of the feedback) would not also yield improved performance? We found for this case indeed an improvement to 88\% (Table~\ref{tab:control}, Augmented FF), which still falls short of the 95\% obtained with the specific feedback in the FB-AlexNet. These controls confirm that feedback based on transfer entropy specifically improves the classification rate.

\begin{table}[h!]
\centering
\caption{Maximal or average (n=100) performance scores for different types of networks.}
{%
\newcommand{\mc}[6]{\multicolumn{#1}{#2}{#3}{#4}{#5}{#6}}
\begin{center}
\begin{tabular}{|c|c|c|c|c|c|}\hline
{FB-AlexNet} & {Shuffle all} & {Shuffle path} & {Uniform weights} & {Uniform Scale} & {Augmented FF}\\\hline\hline
94.62\% & $18.3\pm 7.3$\% & $30.2\pm 10.0$\% & 87.1\% & 85.3\% & 88.0\%\\\hline
\end{tabular}
\end{center}
\label{tab:control}
}%
\end{table}

\subsection*{Discussion}
Here we have shown that feedback defined by transfer entropy between nodes in a small convolutional network can push performance up into the top-level range on the CIFAR benchmark problem, which is otherwise only obtained by much larger networks. The framework employed here is neither limited to this architecture nor to this problem.

Why should one consider using transfer entropy for defining feedback in such networks? Two general observations can be made from the results of this study, which will hold also for other architectures (see Fig.~\ref{fig:TE}). First, there is a decreasing feed-forward convergence towards higher layers, common to most if not all deep convolutional networks. Second, in general transfer entropy is lower between nodes with larger layer distances than between neighbors. This is natural due to fact that long-range transfer entropy is calculated by conditioning on the intermediate layers.

Thus, when using transfer entropy to define feedback, as a consequence there is a higher probability to form long-range as compared to short-range feedback connections in this and in other networks.

If transfer entropy between distant nodes is small then this is indicative of a long path segment, which is "meaningful" and which ought to be more strongly amplified. Thus, as designed here (Eq.~\ref{eqn:fb}), long-range feedback connections should be the strongest. While, on the other hand, pairs of nodes with low transfer entropy in neighboring layers are anyhow directly connected (by their feed-forward link). Their amplification should be small in order to prevent run-away activity.

The framework employed here of using transfer entropy to define staggered feedback naturally serves these requirements: The higher long-range connection probability paired with larger weights counterbalances the reduced forward convergence and visibly amplifies network performance. By contrast, our control experiments show that performance is lower if the system fails to meet these requirements.

The combination of a few-layered network with widespread weak feedback is very reminiscent of many feedback pathways in the vertebrate brain \cite{Gilbert2013,Spillmann2015}. Particularly the interleaving of feed-forward with feedback connectivity discussed by several authors \cite{Markov2013,Markov2014,Yamins2016} suggests that there is also an interleaved information flow happening. It is, thus, tempting to speculate that a similar principle --- an evaluation of the relevance of the different feed-forward pathways --- might have been a phylo- or ontogenetic driving force for the design of different feedback structures in real neural systems. Transfer entropy can be been used to measure how significantly neurons interact \cite{Lizier2011,Vincente2011,Shimono2015}. It is, thus, an interesting question to what degree neural systems might have used this or a similar type of information to distinguish and amplify important relative to less important processing pathways.

\section*{Methods}
\noindent
\underline{\textbf{Defining Transfer Entropies $T_{i \rightarrow j}$:}} The transfer entropy from one neuron (node) $i$ to another upstream neuron $j$ can be understood as the degree of uncertainty reduced in future values of $j$ using knowledge from the past values of $i$ and given past values of $j$. Transfer entropy is, thus, conditional mutual information, using the history of the target variable for conditioning. For time-continuous two-layer systems it is given by:

\begin{align}
  T_{i \rightarrow j}(d) =& \int_{\mathcal{X}_q} \int_{\mathcal{X}_q} \int_{\mathcal{X}_q}
  dy_i(t-d) dy_j(t-d) dy_i(t)  \nonumber \\
  & \cdot p[y_j(t),y_j(t-d),y_i(t-d)] \log \left( \frac{p[y_j(t)|(y_j(t-d),y_i(t-d))]}{p[y_j(t)| y_j(t-d)]} \hfill \right)
 \label{eqn:te1}
 \end{align}
where $p[.]$ is the probability and $d$ defines a reference time point in the past. Intuitively, to calculate $T_{i \rightarrow j}(d)$ the current activity of node $j$ is considered together with the activity of this node as well as that of node $i$ to what they had ''seen'' at time $t-d$ before. Thus: Have these nodes responded to similar inputs in a similar way before? This simplifies for discrete times $t_n$ to:
\begin{equation}
 T_{i\rightarrow j}(d) = \sum_{\mathcal{X}_q} p[y_j(t_{n}), y_j^u(t_{n-d}), y_i^v(t_{n-d})] \log \left( \frac{p[y_j(t_{n})| (y_j^u(t_{n-d}),  y_i^v(t_{n-d}))]}{p[y_j(t_{n})| y_j^u(t_{n-d})]} \right).
 \label{eqn:te2}
\end{equation}
Variables $u$ and $v$ define the summation (integration) window in the past to be used up to $t_{n-d}$. We now set $d=1$ and, thus, consider node activities only for two subsequently analyzed images. We limit summation to this point, too, setting also $u=v=1$ and get:
\begin{equation}
 T_{i\rightarrow j} = \sum_{\mathcal{X}_q} p[y_j(t_{n}), y_j(t_{n-1}), y_i(t_{n-1})] \log \left( \frac{p[y_j(t_{n})| (y_j(t_{n-1}),  y_i(t_{n-1}))]}{p[y_j(t_{n})| y_j(t_{n-1})]} \right).
 \label{eqn:te3}
\end{equation}
To extend this system considering transfer entropies of non-adjacent layers, we have to generalize equation~\ref{eqn:te3} by additionally conditioning it on the nodes in layers between the considered nodes. As we have in the here existing four convolution-layers maximally two more nodes, we get $T_{i \rightarrow j|k}$ or $T_{i \rightarrow j|k|l}$.

%

\bibliographystyle{naturemag-doi}
\bibliographystyle{plain}
\bibliographystyle{unsrt}
\bibliography{general2}

\begin{thebibliography}{10}
\expandafter\ifx\csname url\endcsname\relax
  \def\url#1{\texttt{#1}}\fi
\expandafter\ifx\csname urlprefix\endcsname\relax\def\urlprefix{URL }\fi
\expandafter\ifx\csname doiprefix\endcsname\relax\def\doiprefix{DOI }\fi
\providecommand{\bibinfo}[2]{#2}
\providecommand{\eprint}[2][]{\url{#2}}

\bibitem{inceptionv4}
\bibinfo{author}{Szegedy, C.}, \bibinfo{author}{Ioffe, S.} \&
  \bibinfo{author}{Vanhoucke, V.}
\newblock \bibinfo{title}{Inception-v4, inception-resnet and the impact of
  residual connections on learning}.
\newblock \emph{\bibinfo{journal}{CoRR}}
  \textbf{\bibinfo{volume}{abs/1602.07261}} (\bibinfo{year}{2016}).
\newblock \urlprefix\url{http://arxiv.org/abs/1602.07261}.

\bibitem{resnet}
\bibinfo{author}{He, K.}, \bibinfo{author}{Zhang, X.}, \bibinfo{author}{Ren,
  S.} \& \bibinfo{author}{Sun, J.}
\newblock \bibinfo{title}{Deep residual learning for image recognition}.
\newblock \emph{\bibinfo{journal}{CoRR}}
  \textbf{\bibinfo{volume}{abs/1512.03385}} (\bibinfo{year}{2015}).
\newblock \urlprefix\url{http://arxiv.org/abs/1512.03385}.

\bibitem{Markov2014}
\bibinfo{author}{Markov, N.~T.} \emph{et~al.}
\newblock \bibinfo{title}{{{A}natomy of hierarchy: feedforward and feedback
  pathways in macaque visual cortex}}.
\newblock \emph{\bibinfo{journal}{J. Comp. Neurol.}}
  \textbf{\bibinfo{volume}{522}}, \bibinfo{pages}{225--259}
  (\bibinfo{year}{2014}).

\bibitem{Hermans2013}
\bibinfo{author}{Hermans, M.} \& \bibinfo{author}{Schrauwen, B.}
\newblock \bibinfo{title}{Training and analysing deep recurrent neural
  networks}.
\newblock In \bibinfo{editor}{Burges, C. J.~C.}, \bibinfo{editor}{Bottou, L.},
  \bibinfo{editor}{Welling, M.}, \bibinfo{editor}{Ghahramani, Z.} \&
  \bibinfo{editor}{Weinberger, K.~Q.} (eds.) \emph{\bibinfo{booktitle}{Advances
  in Neural Information Processing Systems 26}}, \bibinfo{pages}{190--198}
  (\bibinfo{publisher}{Curran Associates, Inc.}, \bibinfo{year}{2013}).
\newblock
  \urlprefix\url{http://papers.nips.cc/paper/5166-training-and-analysing-deep-recurrent-neural-networks.pdf}.

\bibitem{Liao2016}
\bibinfo{author}{Liao, Q.} \& \bibinfo{author}{Poggio, T.~A.}
\newblock \bibinfo{title}{Bridging the gaps between residual learning,
  recurrent neural networks and visual cortex}.
\newblock \emph{\bibinfo{journal}{CoRR}}
  \textbf{\bibinfo{volume}{abs/1604.03640}} (\bibinfo{year}{2016}).
\newblock \urlprefix\url{http://arxiv.org/abs/1604.03640}.

\bibitem{hochreiter1997long}
\bibinfo{author}{Hochreiter, S.} \& \bibinfo{author}{Schmidhuber, J.}
\newblock \bibinfo{title}{Long short-term memory}.
\newblock \emph{\bibinfo{journal}{Neural computation}}
  \textbf{\bibinfo{volume}{9}}, \bibinfo{pages}{1735--1780}
  (\bibinfo{year}{1997}).

\bibitem{Salakhutdinov2009}
\bibinfo{author}{Salakhutdinov, R.} \& \bibinfo{author}{Hinton, G.~E.}
\newblock \bibinfo{title}{Deep boltzmann machines}.
\newblock In \emph{\bibinfo{booktitle}{Proceedings of the Twelfth International
  Conference on Artificial Intelligence and Statistics, {AISTATS} 2009,
  Clearwater Beach, Florida, USA, April 16-18, 2009}},
  \bibinfo{pages}{448--455} (\bibinfo{year}{2009}).
\newblock
  \urlprefix\url{http://www.jmlr.org/proceedings/papers/v5/salakhutdinov09a.html}.

\bibitem{Stollenga2014}
\bibinfo{author}{Stollenga, M.~F.}, \bibinfo{author}{Masci, J.},
  \bibinfo{author}{Gomez, F.} \& \bibinfo{author}{Schmidhuber, J.}
\newblock \bibinfo{title}{Deep networks with internal selective attention
  through feedback connections}.
\newblock In \bibinfo{editor}{Ghahramani, Z.}, \bibinfo{editor}{Welling, M.},
  \bibinfo{editor}{Cortes, C.}, \bibinfo{editor}{Lawrence, N.} \&
  \bibinfo{editor}{Weinberger, K.} (eds.) \emph{\bibinfo{booktitle}{Advances in
  Neural Information Processing Systems 27}}, \bibinfo{pages}{3545--3553}
  (\bibinfo{publisher}{Curran Associates, Inc.}, \bibinfo{year}{2014}).
\newblock
  \urlprefix\url{http://papers.nips.cc/paper/5276-deep-networks-with-internal-selective-attention-through-feedback-connections.pdf}.

\bibitem{Cao2015}
\bibinfo{author}{Cao, C.} \emph{et~al.}
\newblock \bibinfo{title}{Look and think twice: Capturing top-down visual
  attention with feedback convolutional neural networks}.
\newblock In \emph{\bibinfo{booktitle}{2015 {IEEE} International Conference on
  Computer Vision, {ICCV} 2015, Santiago, Chile, December 7-13, 2015}},
  \bibinfo{pages}{2956--2964} (\bibinfo{year}{2015}).
\newblock \urlprefix\url{http://dx.doi.org/10.1109/ICCV.2015.338}.

\bibitem{Varadarajan2013}
\bibinfo{author}{Varadarajan, K.~M.} \& \bibinfo{author}{Vincze, M.}
\newblock \bibinfo{title}{Parallel deep learning with suggestive activation for
  object category recognition}.
\newblock In \bibinfo{editor}{Chen, M.}, \bibinfo{editor}{Leibe, B.} \&
  \bibinfo{editor}{Neumann, B.} (eds.) \emph{\bibinfo{booktitle}{Computer
  Vision Systems}}, vol. \bibinfo{volume}{7963} of
  \emph{\bibinfo{series}{Lecture Notes in Computer Science}},
  \bibinfo{pages}{354--363} (\bibinfo{publisher}{Springer},
  \bibinfo{year}{2013}).

\bibitem{Oberweger2015}
\bibinfo{author}{Oberweger, M.}, \bibinfo{author}{Wohlhart, P.} \&
  \bibinfo{author}{Lepetit, V.}
\newblock \bibinfo{title}{Training a feedback loop for hand pose estimation}.
\newblock In \emph{\bibinfo{booktitle}{2015 {IEEE} International Conference on
  Computer Vision, {ICCV} 2015, Santiago, Chile, December 7-13, 2015}},
  \bibinfo{pages}{3316--3324} (\bibinfo{year}{2015}).
\newblock \urlprefix\url{http://dx.doi.org/10.1109/ICCV.2015.379}.

\bibitem{Veeriah2015}
\bibinfo{author}{Veeriah, V.}, \bibinfo{author}{Durvasula, R.} \&
  \bibinfo{author}{Qi, G.-J.}
\newblock \bibinfo{title}{Deep learning architecture with dynamically
  programmed layers for brain connectome prediction}.
\newblock In \emph{\bibinfo{booktitle}{Proceedings of the 21th ACM SIGKDD
  International Conference on Knowledge Discovery and Data Mining}}, KDD '15,
  \bibinfo{pages}{1205--1214} (\bibinfo{publisher}{ACM}, \bibinfo{address}{New
  York, NY, USA}, \bibinfo{year}{2015}).
\newblock \urlprefix\url{http://doi.acm.org/10.1145/2783258.2783399}.

\bibitem{Shi2016}
\bibinfo{author}{Shi, Y.}, \bibinfo{author}{Tian, Y.}, \bibinfo{author}{Wang,
  Y.} \& \bibinfo{author}{Huang, T.}
\newblock \bibinfo{title}{shuttlenet: {A} biologically-inspired {RNN} with loop
  connection and parameter sharing}.
\newblock \emph{\bibinfo{journal}{CoRR}}
  \textbf{\bibinfo{volume}{abs/1611.05216}} (\bibinfo{year}{2016}).
\newblock \urlprefix\url{http://arxiv.org/abs/1611.05216}.

\bibitem{Testolin2016}
\bibinfo{author}{Testolin, A.} \& \bibinfo{author}{Zorzi, M.}
\newblock \bibinfo{title}{Probabilistic models and generative neural networks:
  Towards an unified framework for modeling normal and impaired neurocognitive
  functions}.
\newblock \emph{\bibinfo{journal}{Front. Comput. Neurosci.}}
  \textbf{\bibinfo{volume}{2016}} (\bibinfo{year}{2016}).
\newblock \urlprefix\url{http://dx.doi.org/10.3389/fncom.2016.00073}.
\newblock \doiprefix 10.3389/fncom.2016.00073.

\bibitem{Lillicrap2016}
\bibinfo{author}{Lillicrap, T.~P.}, \bibinfo{author}{Cownden, D.},
  \bibinfo{author}{Tweed, D.~B.} \& \bibinfo{author}{Akerman, C.~J.}
\newblock \bibinfo{title}{{{R}andom synaptic feedback weights support error
  backpropagation for deep learning}}.
\newblock \emph{\bibinfo{journal}{Nat Commun}} \textbf{\bibinfo{volume}{7}},
  \bibinfo{pages}{13276} (\bibinfo{year}{2016}).

\bibitem{Lecun2015}
\bibinfo{author}{LeCun, Y.}, \bibinfo{author}{Bengio, Y.} \&
  \bibinfo{author}{Hinton, G.}
\newblock \bibinfo{title}{{{D}eep learning}}.
\newblock \emph{\bibinfo{journal}{Nature}} \textbf{\bibinfo{volume}{521}},
  \bibinfo{pages}{436--444} (\bibinfo{year}{2015}).

\bibitem{Roudi2015}
\bibinfo{author}{Roudi, Y.} \& \bibinfo{author}{Taylor, G.}
\newblock \bibinfo{title}{{{L}earning with hidden variables}}.
\newblock \emph{\bibinfo{journal}{Curr. Opin. Neurobiol.}}
  \textbf{\bibinfo{volume}{35}}, \bibinfo{pages}{110--118}
  (\bibinfo{year}{2015}).

\bibitem{Khaligh2014}
\bibinfo{author}{Khaligh-Razavi, S.~M.} \& \bibinfo{author}{Kriegeskorte, N.}
\newblock \bibinfo{title}{{{D}eep supervised, but not unsupervised, models may
  explain {I}{T} cortical representation}}.
\newblock \emph{\bibinfo{journal}{PLoS Comput. Biol.}}
  \textbf{\bibinfo{volume}{10}}, \bibinfo{pages}{e1003915}
  (\bibinfo{year}{2014}).

\bibitem{Guclu2015}
\bibinfo{author}{G{\"u}{\c c}l{\"u}, U.} \& \bibinfo{author}{van Gerven, M.
  A.~J.}
\newblock \bibinfo{title}{Deep neural networks reveal a gradient in the
  complexity of neural representations across the ventral stream}.
\newblock \emph{\bibinfo{journal}{Journal of Neuroscience}}
  \textbf{\bibinfo{volume}{35}}, \bibinfo{pages}{10005--10014}
  (\bibinfo{year}{2015}).
\newblock \urlprefix\url{http://www.jneurosci.org/content/35/27/10005}.
\newblock \doiprefix 10.1523/JNEUROSCI.5023-14.2015.
\newblock \eprint{http://www.jneurosci.org/content/35/27/10005.full.pdf}.

\bibitem{Kriegeskorte2015}
\bibinfo{author}{Kriegeskorte, N.}
\newblock \bibinfo{title}{Deep neural networks: a new framework for modelling
  biological vision and brain information processing}.
\newblock \emph{\bibinfo{journal}{Annu. Rev. Vis. Sci.}}
  \textbf{\bibinfo{volume}{1}}, \bibinfo{pages}{417–446}
  (\bibinfo{year}{2015}).
\newblock \doiprefix 10.1146/annurev-vision-082114-035447.

\bibitem{Cichy2016}
\bibinfo{author}{Cichy, R.~M.}, \bibinfo{author}{Khosla, A.},
  \bibinfo{author}{Pantazis, D.}, \bibinfo{author}{Torralba, A.} \&
  \bibinfo{author}{Oliva, A.}
\newblock \bibinfo{title}{Deep neural networks predict hierarchical
  spatio-temporal cortical dynamics of human visual object recognition}.
\newblock \emph{\bibinfo{journal}{CoRR}}
  \textbf{\bibinfo{volume}{abs/1601.02970}} (\bibinfo{year}{2016}).
\newblock \urlprefix\url{http://arxiv.org/abs/1601.02970}.

\bibitem{Yamins2016}
\bibinfo{author}{Yamins, D.~L.} \& \bibinfo{author}{DiCarlo, J.~J.}
\newblock \bibinfo{title}{{{U}sing goal-driven deep learning models to
  understand sensory cortex}}.
\newblock \emph{\bibinfo{journal}{Nat. Neurosci.}}
  \textbf{\bibinfo{volume}{19}}, \bibinfo{pages}{356--365}
  (\bibinfo{year}{2016}).

\bibitem{Bengio2015}
\bibinfo{author}{Bengio, Y.}, \bibinfo{author}{Lee, D.},
  \bibinfo{author}{Bornschein, J.} \& \bibinfo{author}{Lin, Z.}
\newblock \bibinfo{title}{Towards biologically plausible deep learning}.
\newblock \emph{\bibinfo{journal}{CoRR}}
  \textbf{\bibinfo{volume}{abs/1502.04156}} (\bibinfo{year}{2015}).
\newblock \urlprefix\url{http://arxiv.org/abs/1502.04156}.

\bibitem{Marblestone2016}
\bibinfo{author}{Marblestone, A.~H.}, \bibinfo{author}{Wayne, G.} \&
  \bibinfo{author}{Kording, K.~P.}
\newblock \bibinfo{title}{Toward an integration of deep learning and
  neuroscience}.
\newblock \emph{\bibinfo{journal}{Frontiers in Computational Neuroscience}}
  \textbf{\bibinfo{volume}{10}}, \bibinfo{pages}{94} (\bibinfo{year}{2016}).
\newblock
  \urlprefix\url{http://journal.frontiersin.org/article/10.3389/fncom.2016.00094}.
\newblock \doiprefix 10.3389/fncom.2016.00094.

\bibitem{Gilbert2013}
\bibinfo{author}{Gilbert, C.~D.} \& \bibinfo{author}{Li, W.}
\newblock \bibinfo{title}{{{T}op-down influences on visual processing}}.
\newblock \emph{\bibinfo{journal}{Nat. Rev. Neurosci.}}
  \textbf{\bibinfo{volume}{14}}, \bibinfo{pages}{350--363}
  (\bibinfo{year}{2013}).

\bibitem{Spillmann2015}
\bibinfo{author}{Spillmann, L.}, \bibinfo{author}{Dresp-Langley, B.} \&
  \bibinfo{author}{Tseng, C.~H.}
\newblock \bibinfo{title}{{{B}eyond the classical receptive field: {T}he effect
  of contextual stimuli}}.
\newblock \emph{\bibinfo{journal}{J Vis}} \textbf{\bibinfo{volume}{15}},
  \bibinfo{pages}{7} (\bibinfo{year}{2015}).

\bibitem{alexnet}
\bibinfo{author}{Krizhevsky, A.}, \bibinfo{author}{Sutskever, I.} \&
  \bibinfo{author}{Hinton, G.~E.}
\newblock \bibinfo{title}{Imagenet classification with deep convolutional
  neural networks}.
\newblock In \bibinfo{editor}{Pereira, F.}, \bibinfo{editor}{Burges, C. J.~C.},
  \bibinfo{editor}{Bottou, L.} \& \bibinfo{editor}{Weinberger, K.~Q.} (eds.)
  \emph{\bibinfo{booktitle}{Advances in Neural Information Processing Systems
  25}}, \bibinfo{pages}{1097--1105} (\bibinfo{publisher}{Curran Associates,
  Inc.}, \bibinfo{year}{2012}).
\newblock
  \urlprefix\url{http://papers.nips.cc/paper/4824-imagenet-classification-with-deep-convolutional-neural-networks.pdf}.

\bibitem{Lizier2011}
\bibinfo{author}{Lizier, J.~T.}, \bibinfo{author}{Heinzle, J.},
  \bibinfo{author}{Horstmann, A.}, \bibinfo{author}{Haynes, J.~D.} \&
  \bibinfo{author}{Prokopenko, M.}
\newblock \bibinfo{title}{{M}ultivariate information-theoretic measures reveal
  directed information structure and task relevant changes in f{M}{R}{I}
  connectivity}.
\newblock \emph{\bibinfo{journal}{J Comput Neurosci}}
  \textbf{\bibinfo{volume}{30}}, \bibinfo{pages}{85--107}
  (\bibinfo{year}{2011}).

\bibitem{Vincente2011}
\bibinfo{author}{Vicente, R.}, \bibinfo{author}{Wibral, M.},
  \bibinfo{author}{Lindner, M.} \& \bibinfo{author}{Pipa, G.}
\newblock \bibinfo{title}{{T}ransfer entropy--a model-free measure of effective
  connectivity for the neurosciences}.
\newblock \emph{\bibinfo{journal}{J Comput Neurosci}}
  \textbf{\bibinfo{volume}{30}}, \bibinfo{pages}{45--67}
  (\bibinfo{year}{2011}).

\bibitem{Shimono2015}
\bibinfo{author}{Shimono, M.} \& \bibinfo{author}{Beggs, J.~M.}
\newblock \bibinfo{title}{{F}unctional {C}lusters, {H}ubs, and {C}ommunities in
  the {C}ortical {M}icroconnectome}.
\newblock \emph{\bibinfo{journal}{Cereb. Cortex}}
  \textbf{\bibinfo{volume}{25}}, \bibinfo{pages}{3743--3757}
  (\bibinfo{year}{2015}).

\bibitem{Krizhevsky09learningmultiple}
\bibinfo{author}{Krizhevsky, A.}
\newblock \bibinfo{title}{Learning multiple layers of features from tiny
  images}.
\newblock \bibinfo{type}{Tech. Rep.} (\bibinfo{year}{2009}).

\bibitem{Krizhevsky2009}
\bibinfo{author}{Krizhevsky, A.}, \bibinfo{author}{Nair, V.} \&
  \bibinfo{author}{Hinton, G.}
\newblock \bibinfo{title}{Cifar-10 ({C}anadian {I}nstitute for {A}dvanced
  {R}esearch)}.
\newblock \emph{\bibinfo{journal}{Technical Report}}  (\bibinfo{year}{2009}).
\newblock \urlprefix\url{http://www.cs.toronto.edu/~kriz/cifar.html}.

\bibitem{Rumelhart1986}
\bibinfo{author}{{Rumelhart}, D.~E.}, \bibinfo{author}{{Hinton}, G.~E.} \&
  \bibinfo{author}{{Williams}, R.~J.}
\newblock \bibinfo{title}{{Learning representations by back-propagating
  errors}}.
\newblock \emph{\bibinfo{journal}{Nature}} \textbf{\bibinfo{volume}{323}},
  \bibinfo{pages}{533--536} (\bibinfo{year}{1986}).
\newblock \doiprefix 10.1038/323533a0.

\bibitem{Schmidhuber14}
\bibinfo{author}{Schmidhuber, J.}
\newblock \bibinfo{title}{Deep learning in neural networks: An overview}.
\newblock \emph{\bibinfo{journal}{CoRR}}
  \textbf{\bibinfo{volume}{abs/1404.7828}} (\bibinfo{year}{2014}).
\newblock \urlprefix\url{http://arxiv.org/abs/1404.7828}.

\bibitem{Graham14a}
\bibinfo{author}{Graham, B.}
\newblock \bibinfo{title}{Fractional max-pooling}.
\newblock \emph{\bibinfo{journal}{CoRR}}
  \textbf{\bibinfo{volume}{abs/1412.6071}} (\bibinfo{year}{2014}).
\newblock \urlprefix\url{http://arxiv.org/abs/1412.6071}.

\bibitem{springenberg2014striving}
\bibinfo{author}{Springenberg, J.~T.}, \bibinfo{author}{Dosovitskiy, A.},
  \bibinfo{author}{Brox, T.} \& \bibinfo{author}{Riedmiller, M.}
\newblock \bibinfo{title}{Striving for simplicity: The all convolutional net}.
\newblock \emph{\bibinfo{journal}{arXiv preprint arXiv:1412.6806}}
  (\bibinfo{year}{2014}).

\bibitem{MishkinM15}
\bibinfo{author}{Mishkin, D.} \& \bibinfo{author}{Matas, J.}
\newblock \bibinfo{title}{All you need is a good init}.
\newblock \emph{\bibinfo{journal}{CoRR}}
  \textbf{\bibinfo{volume}{abs/1511.06422}} (\bibinfo{year}{2015}).
\newblock \urlprefix\url{http://arxiv.org/abs/1511.06422}.

\bibitem{wibral2014local}
\bibinfo{author}{Wibral, M.}, \bibinfo{author}{Lizier, J.~T.},
  \bibinfo{author}{V{\"o}gler, S.}, \bibinfo{author}{Priesemann, V.} \&
  \bibinfo{author}{Galuske, R.}
\newblock \bibinfo{title}{Local active information storage as a tool to
  understand distributed neural information processing}
  (\bibinfo{year}{2014}).

\bibitem{watts1998collective}
\bibinfo{author}{Watts, D.~J.} \& \bibinfo{author}{Strogatz, S.~H.}
\newblock \bibinfo{title}{Collective dynamics of ‘small-world’networks}.
\newblock \emph{\bibinfo{journal}{Nature}} \textbf{\bibinfo{volume}{393}},
  \bibinfo{pages}{440--442} (\bibinfo{year}{1998}).

\bibitem{rubinov2010complex}
\bibinfo{author}{Rubinov, M.} \& \bibinfo{author}{Sporns, O.}
\newblock \bibinfo{title}{Complex network measures of brain connectivity: uses
  and interpretations}.
\newblock \emph{\bibinfo{journal}{Neuroimage}} \textbf{\bibinfo{volume}{52}},
  \bibinfo{pages}{1059--1069} (\bibinfo{year}{2010}).

\bibitem{latora2001efficient}
\bibinfo{author}{Latora, V.} \& \bibinfo{author}{Marchiori, M.}
\newblock \bibinfo{title}{Efficient behavior of small-world networks}.
\newblock \emph{\bibinfo{journal}{Physical review letters}}
  \textbf{\bibinfo{volume}{87}}, \bibinfo{pages}{198701}
  (\bibinfo{year}{2001}).

\bibitem{Markov2013}
\bibinfo{author}{Markov, N.~T.} \& \bibinfo{author}{Kennedy, H.}
\newblock \bibinfo{title}{The importance of being hierarchical}.
\newblock \emph{\bibinfo{journal}{Current Opinion in Neurobiology}}
  \textbf{\bibinfo{volume}{23}}, \bibinfo{pages}{187 -- 194}
  (\bibinfo{year}{2013}).
\newblock
  \urlprefix\url{http://www.sciencedirect.com/science/article/pii/S0959438813000123}.
\newblock \doiprefix http://doi.org/10.1016/j.conb.2012.12.008.
\newblock \bibinfo{note}{Macrocircuits}.

\end{thebibliography}

\noindent
\underline{\textbf{Acknowledgments}}\\
The research leading to these results has received funding from the European Community’s Horizon 2020 Programme under grant agreement no. 680431, ReconCell. We thank Dr. D. Miner and Dr. M. Tamosiunaite for valuable comments on the manuscript.



\end{document}